\documentclass[12pt]{article}

\usepackage{times,epsfig,graphics,amssymb,latexsym,amsmath,setspace,fullpage,amsthm,subfigure}
\usepackage{array,threeparttable,hyperref,url}
\usepackage[usenames]{color}
\usepackage{graphicx,graphics,psfrag,epsf}
\usepackage{enumerate}
\usepackage{url} 
\usepackage{bm}
\usepackage{hyperref}
\usepackage{multicol}
\usepackage{setspace}
\usepackage{enumerate}
\usepackage{color}
\usepackage[authoryear,sort]{natbib}
\usepackage{footnote}
\usepackage{hyperref}
\usepackage{bigstrut}
\usepackage{booktabs}
\usepackage{xcolor,colortbl}
\usepackage[titletoc]{appendix}
\usepackage{authblk}
\usepackage{filecontents}

\definecolor{lightgray}{gray}{0.9}
\newcolumntype{w}{>{\columncolor{lightgray}}r}
\newcolumntype{u}{>{\columncolor{lightgray}}c}

\def\argmin{\mathop{\rm argmin}}

\newtheorem{Lemma}{Lemma}
\newtheorem{Theorem}{Theorem}

\def\be{\mathop{\bf e}}

\def\trans{^{\rm \textsf{T}}}
\bibpunct[; ]{(}{)}{,}{a}{}{;}

\def\be{\begin{eqnarray}}
\def\bse{\begin{eqnarray*}}
\def\ee{\end{eqnarray}}
\def\ese{\end{eqnarray*}}

\def\wh{\widehat}
\def\wt{\widetilde}
\def\ol{\overline}

\title{On Scalable Inference with Stochastic Gradient Descent}

\author[1]{Yixin Fang}
\author[2]{Jinfeng Xu}
\author[3]{Lei Yang}
\affil[1]{Department of Mathematical Sciences, New Jersey Institute of Technology\thanks{Corresponding to: Cullimore Hall 6th floor, NJIT, Newark, NJ 07102; Email: yixin.fang@njit.edu}}
\affil[2]{Department of Statistics and Actuarial Science , Hong Kong University}
\affil[3]{Department of Population Health, New York University School of Medicine}

\begin{document}

\maketitle


\begin{abstract}
In many applications involving large dataset or online updating, stochastic gradient descent (SGD) provides a scalable way to compute parameter estimates and has gained increasing popularity due to its numerical convenience and memory efficiency. While the asymptotic properties of SGD-based estimators have been established decades ago, statistical inference such as interval estimation remains much unexplored.  The traditional resampling method such as the bootstrap is not computationally feasible since it requires to repeatedly draw independent samples from the entire dataset. The plug-in method is not applicable when there are no explicit formulas for the covariance matrix of the estimator. In this paper, we propose a scalable inferential procedure for stochastic gradient descent, which, upon the arrival of each observation, updates the SGD estimate as well as a large number of randomly perturbed SGD estimates. The proposed method is easy to implement in practice. We establish its theoretical properties for a general class of models that includes generalized linear models and quantile regression models as special cases. The finite-sample performance and numerical utility is evaluated by simulation studies and two real data applications.

\end{abstract}

\bigskip

{Keywords:}  Bootstrap, Interval estimation, Generalized linear models, Large datasets, M-estimators, Quantile regression, Resampling methods, Stochastic gradient descent

\newpage
\setcounter{equation}{0}

\doublespacing

\section{Introduction}
Big datasets arise frequently in clinical, epidemiological, financial and sociological studies. In such applications,
classical optimization methods for parameter estimation such as Fisher scoring, the EM algorithm or iterated reweighted least squares  \citep{hastie2009elements, nelder1972generalized} do not scale well and are computationally less attractive. Due to its computational and memory efficiency, stochastic gradient descent \citep[SGD]{robbins1951stochastic} provides a scalable way for parameter estimation and has recently drawn a great deal of attention. Unlike classical methods that evaluate the objective function involving the entire dataset and require expensive matrix inversions, the SGD method calculates the gradient of the objective function using only one data point at a time and recursively updates the parameter estimate. This is also numerically appealing and particularly useful in online updating settings such as streaming data where it may not even be feasible to retain the entire dataset at the same time. \cite{wang2015statistical} gives a nice review on recent achievements of applying the SGD method to big data and streaming data.

The asymptotic properties of SGD estimators such as consistency and asymptotic normality have been established long time ago; see, for example, \cite{ruppert1988efficient} and \cite{polyak1992acceleration}. However, statistical inference such as confidence interval estimation for SGD estimators has remained largely unexplored. Traditional interval estimation procedures such as the plug-in procedure and the bootstrap are often numerically difficult in the presence of big datasets. The bootstrap repeatedly draws samples from the entire dataset and is thus computationally prohibitive. The plug-in estimator requires an explicit variance-covariance formula and involves expensive matrix inversion. In addition, the bootstrap is not applicable to the online setting where each sample arrives sequentially and it may not be necessary or feasible to store the entire dataset. Neither of them provides a scalable way for interval estimation.

As far as we know, \cite{chen2016statistical} is the only work that considers the statistical inference of the SGD method. Although computationally efficient, their proposed batch-means procedure substantially underestimates the variance of the SGD estimator in finite-sample studies, as shown in the simulation studies of \cite{chen2016statistical}, because of the correlations between the batch means. In addition, the determination of the batch sizes is difficult.

In this paper, we propose a perturbation-based resampling procedure to approximate the distribution of a SGD estimator in a general class of models that include generalized linear models and quantile regression as special cases. Our proposal, justified by asymptotic theories, provides a simple way to estimate the covariance matrix and confidence regions. Through numerical experiments, we verify the ability of this procedure to give accurate inference for big datasets.

The rest of the article is organized as follows. In Section 2,  we introduce the proposed perturbation-based resampling procedure for constructing confidence regions. In Section 3, we theoretically justify the validity of our proposal for a general class of models.  In Section 4, we demonstrate the performance of the proposed procedures in finite samples via simulation studies and two real data applications. Some concluding remarks are given in Section 5 and all the technical proofs are relegated to the Appendix.

\section{The proposed resampling procedure}

Parameter estimation by optimizing an objective function is often encountered  in statistical practice. Consider the general situation where the optimal model parameter $\theta_0\in \mathcal{R}^p$ is defined to be the minimizer of the expected loss function,
\begin{eqnarray}\label{theta-0}
{\theta}_0=\argmin_{\theta}\left\{L(\theta)\triangleq \mathbb{E}[l(\theta; Z)]\right\},
\end{eqnarray}
where $l(\theta; z)$ is some loss function and $Z$ denotes one single observation. Suppose that the data consist of independent and identically distributed (i.i.d.) copies of $Z$,  denoted by $\mathcal{D}_N=\{Z_1, \dots, Z_N\}$. Under mild conditions, $\theta_0$ can be consistently estimated by
 \begin{eqnarray}\label{est-regular}
 \wt{\theta}=\argmin_{\theta}\left\{\frac{1}{N}\sum_{i=1}^{N}l(\theta; Z_i)\right\}.
 \end{eqnarray}
However, the minimization problem (\ref{est-regular}) for big datasets with millions of data points pose numerical challenges for classical methods such as Newton-Raphson algorithm and iteratively reweighted least squares. Furthermore, for applications such as online data where each sample arrives sequentially (e.g., search queries or transactional data), it may not be necessary or feasible to store the entire dataset, leaving alone evaluating the minimand in (\ref{est-regular}).

As a stochastic approximation method \citep{robbins1951stochastic}, stochastic gradient descent provides a scalable way for parameter estimation with large-scale data. Given an initial estimate $\wh{\theta}_0$, the SGD method recursively updates the estimate upon the arrival of each data point $Z_n$,
\begin{eqnarray}\label{SGD}
\wh{\theta}_n=\wh{\theta}_{n-1}-\gamma_n\nabla l(\wh{\theta}_{n-1}; Z_n),
\end{eqnarray}
where $n=1, 2, \dots, N$, and the learning rate $\gamma_n=\gamma n^{-\alpha}$ with $\gamma>0$ and $\alpha\in (0.5, 1)$. As suggested by \cite{ruppert1988efficient} and \cite{polyak1992acceleration}, the final SGD estimate is often taken as the averaging estimate,
\begin{eqnarray}\label{SGD-avg}
\ol{\theta}_N=\frac{1}{N}\sum_{i=1}^N\wh{\theta}_i.
\end{eqnarray}

To order to do statistical inference with the averaging SGD estimator $\ol{\theta}_N$, we propose a perturbation resampling procedure, which recursively updates the SGD estimate as well as a large number of randomly perturbed SGD estimates, upon the arrival of each data point. Specifically, let $\mathcal{W} = \{W_i, i = 1, \dots, N\}$ be a set of i.i.d. non-negative random variables with mean and variance equal to one. In parallel with (\ref{SGD}) and (\ref{SGD-avg}), with $\wh{\theta}^*_0\equiv \wh{\theta}_0$, upon observing data point $Z_n$, we recursively updates randomly perturbed SGD estimates,
\begin{eqnarray}
\wh{\theta}^*_n&=&\wh{\theta}^*_{n-1}-\gamma_nW_n\nabla l(\wh{\theta}^*_{n-1}; Z_n),\label{SGD-wt}\\
\ol{\theta}^*_n&=&\frac{1}{n}\sum_{i=1}^n\wh{\theta}^*_i.\label{SGD-avg-wt}
\end{eqnarray}

We will show that $\sqrt{n}(\ol{\theta}_n-{\theta}_0)$ and $\sqrt{n}(\ol{\theta}^*_n-\ol{\theta}_n)$ converge in distribution to the same limiting distribution. In practice, these results allow us to estimate the distribution of $\sqrt{n}(\ol{\theta}_n-{\theta}_0)$ by generating a large number, say $B$, of random samples of $\mathcal{W}$. We obtain  $\ol{\theta}^{*,b}_n$ by sequentially updating perturbed SGD estimates for each sample, $b=1, \dots, B$,
 \begin{eqnarray}
 \wh{\theta}^{*,b}_n&=&\wh{\theta}^{*,b}_{n-1}-\gamma_nW_{n,b}\nabla l(\wh{\theta}^{*,b}_{n-1}; Z_n),\label{SGD-wt-b}\\
 \ol{\theta}^{*,b}_n&=&\frac{1}{n}\sum_{i=1}^n\wh{\theta}^{*,b}_i,\label{SGD-avg-wt-b}
 \end{eqnarray}
and then approximate the sampling distribution of $\ol{\theta}_n$ by the empirical distribution of $\{\ol{\theta}^{*,b}_n,b = 1,...,B\}$. Specifically, the covariance matrix of $\ol{\theta}_n$ can be estimated by the sample covariance matrix constructed from $\{\ol{\theta}^{*,b}_n,b = 1,...,B\}$. Estimating the distribution of $\sqrt{n}(\ol{\theta}_n-{\theta}_0)$ based on the distribution of $\sqrt{n}(\ol{\theta}^*_n-\ol{\theta}_n)|\mathcal{D}_n$ leads to the construction of $(1-\alpha)100\%$ confidence regions for $\theta_0$. The resulting inferential procedure retains the numerical simplicity of the SGD method, only using one pass over the data. The proposed inferential procedure scales well for datasets with millions of data points and the theoretical validity can be justified under two general model settings with mild regularity conditions as shown in the next section.

\section{Theoretical Results}

In this section, we derive the theoretical properties of $\ol{\theta}_n^*$, justifying that the conditional distribution of $\ol{\theta}_n^*$ given data $\mathcal{D}_n=\{Z_1, Z_2, \dots, Z_n\}$ can approximate the sampling distribution of $\ol{\theta}_n$, under the following two model settings.

\subsection{Model Setting 1}

We first consider the setting where the objective function, $L(\theta)$ in (\ref{theta-0}), is smooth. This includes linear regression, logistic regression and other generalized linear models as special cases. To ensure the consistency and asymptotic properties of the SGD estimator and the validity of the proposed resampling procedure, we assume the following assumptions.
\begin{enumerate}
\item[(A1).] The objective function $L(\theta)$ is continuously differentiable and strongly convex with constant $\lambda>0$; that is, for any $\theta_1$ and $\theta_2$, $L(\theta_2)\geq L(\theta_1)+[\nabla L(\theta_1)]\trans(\theta_2-\theta_1)+\lambda\|\theta_2-\theta_1\|_2^2$.

\item[(A2).] The gradient of $L(\theta)$, $\nabla L(\theta)$, is Lipchitz continuous with constant $L_0>0$; that is, for any  $\theta_1$ and $\theta_2$, $\|\nabla L(\theta_1)-\nabla L(\theta_2)\|_2\leq L_0\|\theta_1-\theta_2\|_2$.

\item[(A3).] Let $S(\theta)=\nabla^2 L(\theta)$ be the Hessian matrix of $L(\theta)$. Assume that $S(\theta)$ exists and is continuous in a neighborhood of $\theta_0$. And assume that $S=S(\theta_0)>0$.

\item[(A4).] Let $V=\mathbb{E}\left\{[\nabla l(\theta_0; Z)][\nabla l(\theta_0; Z)]\trans\right\}$.  Let $v(\theta)=\mathbb{E}\left\{\|\nabla l(\theta; Z)\|_2^2\right\}$ and assume $v(\theta)\leq C(1+\|\theta\|^2_2)$ for some $C>0$. Assume  $\mathbb{E}\left\{\|\nabla l(\theta; Z)-\nabla l(\theta_0; Z)\|^2_2\right\}\rightarrow 0$ as $\theta \rightarrow \theta_0$.
\end{enumerate}

Following similar arguments in  \cite{ruppert1988efficient} and \cite{polyak1992acceleration}, the SGD estimator $\ol{\theta}_n$ is asymptotically normal under Model Setting 1.
{\Lemma If Assumptions A1-A4 are satisfied, then we have
\begin{eqnarray}\label{normality-set1}
\sqrt{n}(\ol{\theta}_n-\theta_0) \Rightarrow \mathcal{N}\left(0, S^{-1}VS^{-1}\right), {\mbox \ in \ distribution\ as\ } n \rightarrow \infty.
\end{eqnarray}
}

By Lemma 1, we can use the plug-in procedure to estimate the asymptotic covariance matrix of $\ol{\theta}_n$, where $S$ and $V$ can be conveniently estimated recursively using
\begin{eqnarray}
\wh{S}_n&=&\frac{1}{n}\sum_{i=1}^n \nabla^2 l(\wh{\theta}_i; Z_i), \label{plug-in-1}\\
\wh{V}_n&=&\frac{1}{n}\sum_{i=1}^n [\nabla l(\wh{\theta}_i; Z_i)][\nabla l(\wh{\theta}_i; Z_i)]\trans. \label{plug-in-2}
\end{eqnarray}
We illustrate this setting and the regularity conditions in two examples. The data consist of $Z_n=(Y_n, X_n)$, $n=1, 2, \dots,$ which are i.i.d. as $Z=(Y,X)$, where $Y_n$ denotes the response variable and $X_n$ be the $p$-dimensional vector of covariates. Assume that $E\|X\|_2^3<\infty$.

{\bf Example 1} (Linear regression) Suppose that $Z_n=(Y_n, X_n)$, $n=1, 2, \dots$, are from the linear regression model,
\begin{eqnarray}\label{eg-lm}
Y_n=X_n\trans\theta_0+\varepsilon_n.
\end{eqnarray}
Assume $\varepsilon_n$ are i.i.d.~with $\varepsilon$, and that $\varepsilon$ and $X$  are mutually independent and $E\varepsilon^2<\infty$.  Let $l(\theta; Z)=(Y-X\trans\theta)^2$, $\nabla l(\theta; Z)=-2(Y-X\trans\theta)X$, and $\nabla L(\theta)=\mathbb{E}\{\nabla l(\theta; Z)\}=2\mathbb{E}\{XX\trans\}\theta-2\mathbb{E}\{XY\}$. It can
 be easily verified that Assumptions A1-A4 hold and the SGD and perturbed SGD updates for $\theta_0$, as defined in (\ref{SGD}) and (\ref{SGD-wt}) respectively, are
\begin{eqnarray}
\wh{\theta}_n&=&\wh{\theta}_{n-1}+2\gamma_n(Y_n-X_n\trans\wh{\theta}_{n-1})X_n,\label{eg1-sgd}\\
\wh{\theta}^*_n&=&\wh{\theta}^*_{n-1}+2\gamma_nW_n(Y_n-X_n\trans\wh{\theta}^*_{n-1})X_n.\label{eg1-sgd-rw}
\end{eqnarray}

{\bf Example 2} (Logistic regression) Suppose that $Z_n=(Y_n, X_n)$, $n=1, 2, \dots,$ are from the logistic regression model,
\begin{eqnarray}\label{eg-logit}
\mathbb{P}(Y_n=1|X_n)=1-P(Y_n=-1|X_n)=\frac{\exp(X_n\trans\theta_0)}{1+\exp(X_n\trans\theta_0)}.
\end{eqnarray}
Let $l(\theta; Z)=\log\left(1+\exp(-YX\trans\theta)\right)$, $\nabla l(\theta; Z)=-XY/[1+\exp(YX\trans\theta)]$, and $\nabla L(\theta)=\mathbb{E}\left\{{X(e^{X\trans\theta}-e^{X\trans\theta_0})}/{[(1+e^{X\trans\theta})(1+e^{X\trans\theta_0})]}\right\}$. It can be verified that Assumptions A1-A4 hold. The SGD and perturbed SGD updates for $\theta_0$, as defined in (\ref{SGD}) and (\ref{SGD-wt}) respectively, are
\begin{eqnarray}
\wh{\theta}_n&=&\wh{\theta}_{n-1}+\gamma_nXY/[1+\exp(YX\trans\wh{\theta}_{n-1})],\label{eg2-sgd}\\
\wh{\theta}^*_n&=&\wh{\theta}^*_{n-1}+\gamma_nW_nXY/[1+\exp(YX\trans\wh{\theta}^*_{n-1})].\label{eg2-sgd-rw}
\end{eqnarray}

\subsection{Model Setting 2}
The model setting 1 includes smooth objective function in general, not necessarily restricted to the regression case.
Next we consider a general regression setting that allows for non-smooth loss functions, including
quantile regression as a special case. Suppose that the data, $Z_n=(Y_n, X_n)$, $n=1, 2, \dots$, are from the model (\ref{eg-lm}),
and the loss  function is
\begin{equation}\label{M-est}
l(\theta; Z_n)=\rho(Y_n-X_n\trans\theta),
\end{equation}
where $\rho(u)$ is a convex function with $\rho(0)=0$.  We require the following regularity conditions.

\begin{enumerate}
\item[(B1).] Assume that $\{(X_n,\varepsilon_n), n=1,2,...\}$  are i.i.d. copies of
$(X,\varepsilon)$, $X$ and $\varepsilon$ are mutually independent,   $E\|X\|^4_2<\infty$ and  $E\|\varepsilon\|_2^2<\infty$.  Let $G=\mathbb{E}\{XX\trans\}>0$.

\item[(B2).] Assume that $\rho(u)$ is a convex function on $\mathcal{R}$ with the right derivative being $\psi_+(u)$ and left derivative being $\psi_-(u)$. Let $\psi(u)$ be a function such that $\psi_+(u)\leq \psi(u) \leq \psi_-(u)$. There exists constant $C_1>0$ such that $|\psi(u)|\leq C_1(1+|u|)$.

\item[(B3).] Let $\phi(u)=\mathbb{E}\{\psi(u+\varepsilon)\}$. Assume that $\phi(0)=0$, $u\phi(u)>0$ for any $u\neq 0$, and $\phi(u)$ has a derivative at $u=0$ with $\dot{\phi}(0)>0$. There exist constants $C_2>0$ and $\delta>0$ such that $|\phi(u)-\dot{\phi}(0)u|\leq C_2u^2$ for $|u|\leq \delta$.

\item[(B4).] Let $\varphi(u)=\mathbb{E}\{\psi^2(u+\varepsilon)\}$. Assume that $\varphi(u)$ is finite for $u$ in a neighborhood of $u=0$ and is continuous at $u=0$.

\end{enumerate}

By Assumption B2, the SGD and perturbed SGD updates for $\theta_0$, as defined in (\ref{SGD}) and (\ref{SGD-wt}) respectively, are
\begin{eqnarray}
\wh{\theta}_n&=&\wh{\theta}_{n-1}+\gamma_n\psi(Y_n-X_n\trans\wh{\theta}_{n-1})X_n,\label{SGD-M}\\
\wh{\theta}^*_n&=&\wh{\theta}^*_{n-1}+\gamma_nW_n\psi(Y_n-X_n\trans\wh{\theta}^*_{n-1})X_n.\label{SGD-M-wt}
\end{eqnarray}
We establish the asymptotic normality of SGD estimator under model setting 2 as follows.
{\Lemma If Assumptions B1-B4 are satisfied, then we have
\begin{eqnarray}\label{normality-set2}
\sqrt{n}(\ol{\theta}_n-\theta_0) \Rightarrow \mathcal{N}\left(0, G^{-1}{\varphi(0)}/{\dot{\phi}^2(0)}\right), {\mbox \ in \ distribution.}
\end{eqnarray}}

We illustrate the model setting 2 with two examples.

{\bf Example 1} (Linear regression). We revisit  Example 1. Let $\rho(u)=u^2$. We have $\psi(u)=2u$, $\phi(u)=2u+2\mathbb{E}\{\varepsilon\}$, and $\varphi(u)=4\mathbb{E}\{(u+\varepsilon)^2\}$. Thus, $\phi(0)=0$ is equivalent to $\mathbb{E}\{\varepsilon\}=0$, and consequently $\phi(u)=2u$ and $\dot{\phi}(0)=2$. In addition, $\varphi(0)=4\mathbb{E}\{(u+\varepsilon)^2\}=4\sigma^2$. Therefore, the asymptotic covariance matrix in (\ref{normality-set2}) is $\sigma^2 G^{-1}$.

{\bf Example 3} (Quantile regression). Consider $\rho_\tau(u)=u(\tau-I(u<0))$, where $0<\tau<1$. Then $\psi(u)=\tau-I(u<0)$, $\phi(u)=\tau-P(u+\varepsilon<0)$, and $\varphi(u)=\tau(1-\tau)$. Thus, $\phi(0)=0$ is equivalent to that the $\tau$-quantile  of $\varepsilon$ is 0, and  $\dot{\phi}(0)=p_{\varepsilon}(0)$, where $p_{\varepsilon}(u)$ is the density of $\varepsilon$. Then the SGD and perturbed SGD updates for $\theta_0$, as defined in (\ref{SGD}) and (\ref{SGD-wt}) respectively, are
\begin{eqnarray}
\wh{\theta}_n&=&\wh{\theta}_{n-1}+\gamma_n\left\{\tau-I(Y_n-X_n\trans\wh{\theta}_{n-1}<0)\right\}X_n,\label{eg3-sgd}\\
\wh{\theta}^*_n&=&\wh{\theta}^*_{n-1}+\gamma_nW_n\left\{\tau-I(Y_n-X_n\trans\wh{\theta}^*_{n-1}<0)\right\}X_n,\label{eg3-sgd-rw}
\end{eqnarray}
and the asymptotic covariance matrix in (\ref{normality-set2}) is $G^{-1}\tau(1-\tau)/[p^2_{\varepsilon}(0)]$.
As the covariance matrix involves the unknown density function, the
plug-in procedure is not applicable in this example.

\subsection{Asymptotic properties}

Let $\mathbb{P}^*$ and $\mathbb{E}^*$ denote the conditional probability and expectation given the data $\mathcal{D}_n$, respectively. Note that
the perturbation variables $W_1, W_2, \dots$ satisfying that $\mathbb{E}\{W_n\}=\mbox{Var}(W_n)=1$ and
the learning rate $\gamma_i=\gamma n^{-\alpha}$ with $\gamma>0$ and $\alpha\in(0.5, 1)$. We derive the following two theorems for Modeling Setting 1 and 2 respectively.

{\Theorem (Model Setting 1) If Assumptions A1-A4 hold, then we have (i),
\begin{equation}\label{eq-th1}
\sqrt{n}(\ol{\theta}_n^*-\theta_0)=-\frac{1}{\sqrt{n}}S^{-1}\sum_{i=1}^n W_i \nabla l(\theta_0; Z_i)+o_p(1),
\end{equation}
and (ii),
\begin{equation}\label{eq-co}
\sup_{v\in\mathcal{R}^p}\left|\mathbb{P}^*\left(\sqrt{n}(\ol{\theta}^*_n-\ol{\theta}_n)\leq v\right)-\mathbb{P}\Big(\sqrt{n}(\ol{\theta}_n-\theta_0)\leq v\Big)\right| \rightarrow 0, {in\ probability. }
\end{equation}
}

{\Theorem (Model setting 2) If Assumptions B1-B4 hold,  then we have (i),
\begin{equation}\label{eq-th1}
\sqrt{n}(\ol{\theta}_n^*-\theta_0)=\frac{1}{\sqrt{n}\dot{\phi}(0)}G^{-1}\sum_{i=1}^n W_i \psi(\varepsilon_i)X_n+o_p(1),
\end{equation}
and (ii),
\begin{equation}\label{eq-co}
\sup_{v\in\mathcal{R}^p}\left|\mathbb{P}^*\left(\sqrt{n}(\ol{\theta}^*_n-\ol{\theta}_n)\leq v\right)-\mathbb{P}\Big(\sqrt{n}(\ol{\theta}_n-\theta_0)\leq v\Big)\right| \rightarrow 0, { in\ probability. }
\end{equation}
}

By Theorem 1 and 2, under either Modeling Setting 1 or Model Setting 2, the Kolmogorow-Smirnov distance between $\sqrt{n}(\ol{\theta}^*_n-\ol{\theta}_n)$ and $\sqrt{n}(\ol{\theta}_n-\theta_0)$ converges to zero in probability. This validates our proposal of the perturbation-based resampling procedure for inference with SGD.

\section{Numerical results}

\subsection{Simulation studies}

To assess the performance of the proposed perturbation-based resampling (a.k.a. random weighting; RW) procedure for SGD estimators, we conduct simulation studies for those three examples discussed in Section 3. We compare the proposed procedure with the plug-in procedure, if applicable, as described in (\ref{plug-in-1}) and (\ref{plug-in-2}). We don't compare the batch-means procedure proposed by \cite{chen2016statistical}, because their program is not available to public and depends on several tunings (personal communications).

{\bf Example 1} (Least-squares regression):  Consider model (\ref{eg-lm}), where covariates $X^{(j)}$ and error $\varepsilon$ are independently generated from standard normal $N(0,1)$. Here $X^{(j)}$ indicates the $j$-th dimension of $X$. Let $\theta_0=(\mu{\bf 1}\trans_{q/2},-\mu{\bf 1}\trans_{q/2},{\bf 0}\trans_{p-q})\trans$ (same for the other two examples). Consider least-squares (LS) regression and the corresponding SGD estimators are the ones defined in (\ref{eg1-sgd}) and (\ref{eg1-sgd-rw}).

{\bf Example 2} (Logistic regression): Consider logistic (Logit) regression (\ref{eg-logit}), where covariates $X^{(j)}$ are independently generated from $N(0,1)$ and response $Y$ is generated from Bernoulli distribution. The corresponding SGD estimators are the ones defined in (\ref{eg2-sgd}) and (\ref{eg2-sgd-rw}).

{\bf Example 3} (Least-absolute-deviation regression): Consider model (\ref{eg-lm}), where covariates $X^{(j)}$ and error $\varepsilon$ are independently generated from $N(0, 1)$ and $DE(0, 1)$ respectively. Consider quantile regression with $\tau=1/2$, which is equivalent to least-absolute-deviation (LAD) regression. The corresponding SGD estimators are the ones defined in (\ref{eg3-sgd}) and (\ref{eg3-sgd-rw}) with $\tau=1/2$.

%

\begin{table}[!h]
	\centering
	\caption{Coverage probabilities of 95\% confidence intervals for LS regression. }
	\medskip
	\label{tab:sim1}
	\begin{tabular}{ccccc}
        \hline
		$(N, p, q, \mu)$ & Method &Dim 1& Dim $q/2+1$ & Dim $q+1$\\
		\hline\hline
		(10000,10,6,0.1) & RW & 0.962& 0.946 & 0.948\\
		&Plug in & 0.901& 0.917& 0.900\\
		\hline
		(10000,10,6,0.2) & RW & 0.940& 0.948& 0.953\\
		&Plug in & 0.898& 0.924& 0.902\\
		\hline
		(10000,10,6,0.3) & RW  & 0.937& 0.945& 0.943\\
		&Plug in & 0.908& 0.904& 0.906\\
		\hline
		(20000,20,6,0.1) & RW & 0.952& 0.966& 0.969\\
		&Plug in & 0.893& 0.882& 0.902\\
		\hline
		(20000,20,6,0.2)& RW & 0.957& 0.962& 0.969\\
		&Plug in & 0.918& 0.902& 0.927\\
		\hline
		(20000,20,6,0.3) & RW &0.965&0.954&0.961\\
		&Plug in & 0.913& 0.918& 0.926\\
		\hline
	\end{tabular}
\end{table}

\begin{table}[!h]
	\centering
	\caption{Averaged estimated SE and empirical SE for LS regression.}
	\medskip
	\label{tab:sim1-1}
	\begin{tabular}{ccccc}
		\hline
		$(N, p, q, \mu)$ & Method &Dim 1& Dim $q/2+1$ & Dim $q+1$ \\
		\hline\hline
		(10000,10,6,0.1) & RW & 0.0157& 0.0158 & 0.0158 \\
		&Plug in & 0.0137& 0.0137& 0.0137\\
		&Empirical& 0.0156&  0.0157& 0.0158\\
		\hline
		(10000,10,6,0.2) &  RW & 0.0158&0.0158 & 0.0158 \\
		&Plug in & 0.0137& 0.0137& 0.0137\\
		&Empirical& 0.0164&  0.0157& 0.0154 \\
		\hline
		(10000,10,6,0.3) & RW & 0.0158&0.0158 & 0.0158 \\
		&Plug in & 0.0137& 0.0137& 0.0137\\
		&Empirical& 0.0164&  0.0162& 0.0163 \\
		\hline
		(20000,20,6,0.1) & RW & 0.0114&  0.0114& 0.0114 \\
		&Plug in & 0.0096& 0.0096&  0.0096\\
		&Empirical&  0.0114&   0.0104&0.0104  \\
		\hline
		(20000,20,6,0.2)& RW  & 0.0114&  0.0114& 0.0115 \\
		&Plug in & 0.0096& 0.0096&  0.0096\\
		&Empirical& 0.0109&  0.0108& 0.0105 \\
		\hline
		(20000,20,6,0.3) & RW & 0.0115&  0.0115& 0.0114 \\
		&Plug in & 0.0096& 0.0096&  0.0096\\
		&Empirical& 0.0108&  0.0108& 0.0107 \\
		\hline
	\end{tabular}
\end{table}

\begin{table}[!h]
	\centering
	\caption{Coverage probabilities of 95\% confidence intervals for Logit regression. }
	\medskip
	\label{tab:sim2}
	\begin{tabular}{ccccc}
		\hline
		$(N, p, q, \mu)$ & Method &Dim 1& Dim $q/2+1$ & Dim $q+1$\\
		\hline\hline
		(10000,10,6,0.1)  & RW & 0.955& 0.955& 0.961\\
		&Plug in & 0.900& 0.910& 0.877\\
		\hline
		(10000,10,6,0.2) & RW & 0.969& 0.951& 0.956\\
		&Plug in & 0.887& 0.878& 0.878\\
		\hline
		(10000,10,6,0.3) & RW & 0.966&  0.970& 0.954\\
		&Plug in & 0.881& 0.886& 0.895\\
		\hline
		(20000,20,6,0.1) & RW & 0.947& 0.960& 0.943\\
		&Plug in & 0.878& 0.891& 0.890\\
		\hline
		(20000,20,6,0.2)& RW & 0.957& 0.952& 0.931\\
		&Plug in & 0.891& 0.875& 0.885\\
		\hline
		(20000,20,6,0.3) & RW & 0.963& 0.959& 0.938\\
		&Plug in & 0.861& 0.853& 0.861\\
		\hline
	\end{tabular}
\end{table}

\begin{table}[!h]
	\centering
	\caption{Averaged estimated SE and empirical SE for Logit regression.}
	\medskip
	\label{tab:sim2-1}
	\begin{tabular}{ccccc}
		\hline
		$(N, p, q, \mu)$ & Method &Dim 1& Dim $q/2+1$ & Dim $q+1$\\
		\hline\hline
		(10000,10,6,0.1) & RW &0.0234 &  0.0233& 0.0232\\
		&Plug in & 0.0119&  0.0119& 0.0120\\
		&Empirical& 0.0228& 0.0227 & 0.0231\\
		\hline
		(10000,10,6,0.2) &  RW & 0.0246&  0.0246& 0.0240\\
		&Plug in & 0.0111&  0.0111& 0.0111\\
		&Empirical& 0.0226&  0.0241& 0.0229\\
		\hline
		(10000,10,6,0.3) & RW & 0.0268&  0.0268& 0.0254\\
		&Plug in & 0.0100& 0.0100& 0.0103\\
		&Empirical& 0.0250&  0.0245& 0.0251\\
		\hline
		(20000,20,6,0.1) & RW & 0.0158&  0.0157& 0.0157\\
		&Plug in & 0.0084& 0.0084 & 0.0085\\
		&Empirical& 0.0160&  0.0153& 0.0156\\
		\hline
		(20000,20,6,0.2)& RW & 0.0165& 0.0165 & 0.0161\\
		&Plug in & 0.0078&  0.0078& 0.0079\\
		&Empirical& 0.0161&  0.0164& 0.0168\\
		\hline
		(20000,20,6,0.3) & RW &0.0182 & 0.0181 & 0.0169\\
		&Plug in & 0.0069&  0.0069& 0.0073\\
		&Empirical& 0.0166&  0.0173& 0.0170\\
		\hline
	\end{tabular}
\end{table}

\begin{table}[!h]
	\centering
	\caption{Coverage probabilities for 95\% confidence intervals for LAD regression. }
	\medskip
	\label{tab:sim3}
	\begin{tabular}{ccccc}
		\hline
		$(N, p, q, \mu)$ & Method &Dim 1& Dim $q/2+1$ & Dim $q+1$\\
		\hline\hline
		(10000,10,6,0.1) & RW & 0.968& 0.956& 0.960\\
		&Plug in & $-$& $-$& $-$\\
		\hline
		(10000,10,6,0.2) & RW & 0.958& 0.953& 0.966\\
		&Plug in & $-$& $-$& $-$\\
		\hline
		(10000,10,6,0.3) & RW & 0.956& 0.963& 0.959\\
		&Plug in & $-$& $-$& $-$\\
		\hline
		(20000,20,6,0.1) & RW & 0.971& 0.962& 0.969\\
		&Plug in & $-$& $-$& $-$\\
		\hline
		(20000,20,6,0.2)& RW & 0.959& 0.969&0.966\\
		&Plug in & $-$& $-$& $-$\\
		\hline
		(20000,20,6,0.3) & RW & 0.953& 0.959& 0.960\\
		&Plug in & $-$& $-$& $-$\\
		\hline
	\end{tabular}
\end{table}

\begin{table}[!h]
	\centering
	\caption{Averaged estimated SE and empirical SE for LAD regression.}
	\medskip
	\label{tab:sim3-1}
	\begin{tabular}{ccccc}
		\hline
		$(N, p, q, \mu)$ & Method &Dim 1& Dim $q/2+1$ & Dim $q+1$\\
		\hline\hline
		(10000,10,6,0.1) & RW & 0.0130&  0.0129& 0.0129\\
		&Plug in & $-$& $-$& $-$\\
		&Empirical& 0.0119&  0.0117& 0.0120\\
		\hline
		(10000,10,6,0.2) &  RW & 0.0130&  0.0129& 0.0129\\
		&Plug in & $-$& $-$& $-$\\
		&Empirical& 0.0121&  0.0120& 0.0120\\
		\hline
		(10000,10,6,0.3) & RW & 0.0129&  0.0130& 0.0130\\
		&Plug in & $-$& $-$& $-$\\
		&Empirical& 0.0129&  0.0117& 0.0122\\
		\hline
		(20000,20,6,0.1) & RW & 0.0091&  0.0090& 0.0091\\
		&Plug in & $-$& $-$& $-$\\
		&Empirical& 0.0081&  0.0085& 0.0081\\
		\hline
		(20000,20,6,0.2)& RW & 0.0090&  0.0090& 0.0090\\
		&Plug in & $-$& $-$& $-$\\
		&Empirical& 0.0086& 0.0081 & 0.0083\\
		\hline
		(20000,20,6,0.3) & RW & 0.0091&  0.0090& 0.0090\\
		&Plug in & $-$& $-$& $-$\\
		&Empirical& 0.0083&  0.0083&  0.0084\\
		\hline
	\end{tabular}
\end{table}

For each example, we consider six scenarios, as described by $(N, p, q, \mu)$, where sample size $N=10000$ or $20000$, number of covaraites $p=10$ or $20$, number of useful covariates $q=6$, and effect size $\mu=0.1, 0.2$ or $0.3$. For each example, we repeat the data generation 1000 times. For each data repetition, we use $W_{nb}\sim{\rm exp}(1)$ as random weights and generate $B=200$ copies of random weights whenever a new data point is read. Then, for each data repetition, we obtain the SGD estimator (\ref{SGD-avg}), apply the proposed perturbation-based resampling procedure to estimate its standard error, and apply the plug-in procedure (if applicable) to estimate its standard error as well. When we calculate the average SGD estimators (\ref{SGD-avg}) and (\ref{SGD-avg-wt}), the first 2000 estimates are excluded. Based on the estimated standard error $\wh{SE}$, we can construct 95\% confidence interval estimate with form of $\wh{\theta} \pm 1.96 \times \wh{SE}$ and see if it covers the true estimand. We also obtain the empirical standard error based on 1000 repeated SGD estimators, which are considered as a good approximation to the true standard error.

The coverage probabilities of the 95\% confidence interval estimates constructed using our procedure (RW) and the plug-in procedure (Plug-in) are summarized in Tables \ref{tab:sim1},  \ref{tab:sim2} and \ref{tab:sim3} for Examples 1-3 respectively. We only report results corresponding to the first, fourth and seventh covariates and the plug-in procedure is not applicable for Example 3. From these tables, we see that the coverage probabilities from the RW procedure are close to 95\%, while those from the plug-in procedures are substantially smaller than 95\%. Similar findings of the plug-in procedure were also reported in \cite{chen2016statistical}. Therefore, our procedure outperforms the plug-in procedure.

We also compare the average estimated standard errors (SE) using the RW and plug-in procedures with those empirical standard errors, which are thought to be close to the true standard error. The results are summarized in Tables  \ref{tab:sim1-1},  \ref{tab:sim2-1} and \ref{tab:sim3-1} for Examples 1-3 respectively. Again, we only report results corresponding to those three covariates and the plug-in procedure is not applicable for Example 3. From these tables, we see that the average estimated standard errors using the RW procedure are close to those empirical standard errors, while the average estimated standard errors from the plug-in procedure are substantially smaller.

%


\subsection{Real data applications}

In this section, we apply the proposed method to conduct linear regression analysis for the individual household electric power consumption dataset (POWER) and logistic regression analysis for the gas sensors for home activity monitoring dataset (GAS). Both the POWER data and the GAS data are publicly available on UCI machine learning repository.

The POWER data contains 2,075,259 observations and we fit linear regression model to investigate the relationship between the time and response variable ``sub-metering-1", the energy sub-metering No. 1, in watt-hour of active energy, which corresponds to the kitchen, containing mainly a dishwasher, an oven and a microwave. The observations with missing value are deleted and the time are divided into 8 categories, including ``0-2", ``3-5",``6-8", ``9-11", ``12-14", ``15-17", ``18-20" and ``21-23". The GAS data  constains 919,438 observations and we only use a subset containing 652,024 observations with response value being either ``banana" or ``wine". We consider logistic regression model to examine the association between the response variable and 11 covariates, including {\it time}, {\it R1} to {\it R8}, {\it temperature} and {\it humidity}.

\begin{figure}[!h]
	\centering
	\caption{Histograms of $B=1000$ perturbation-based SGD estimates for the POWER data.}
	\includegraphics[height=0.6\textwidth,width=0.8\textwidth,angle=0] {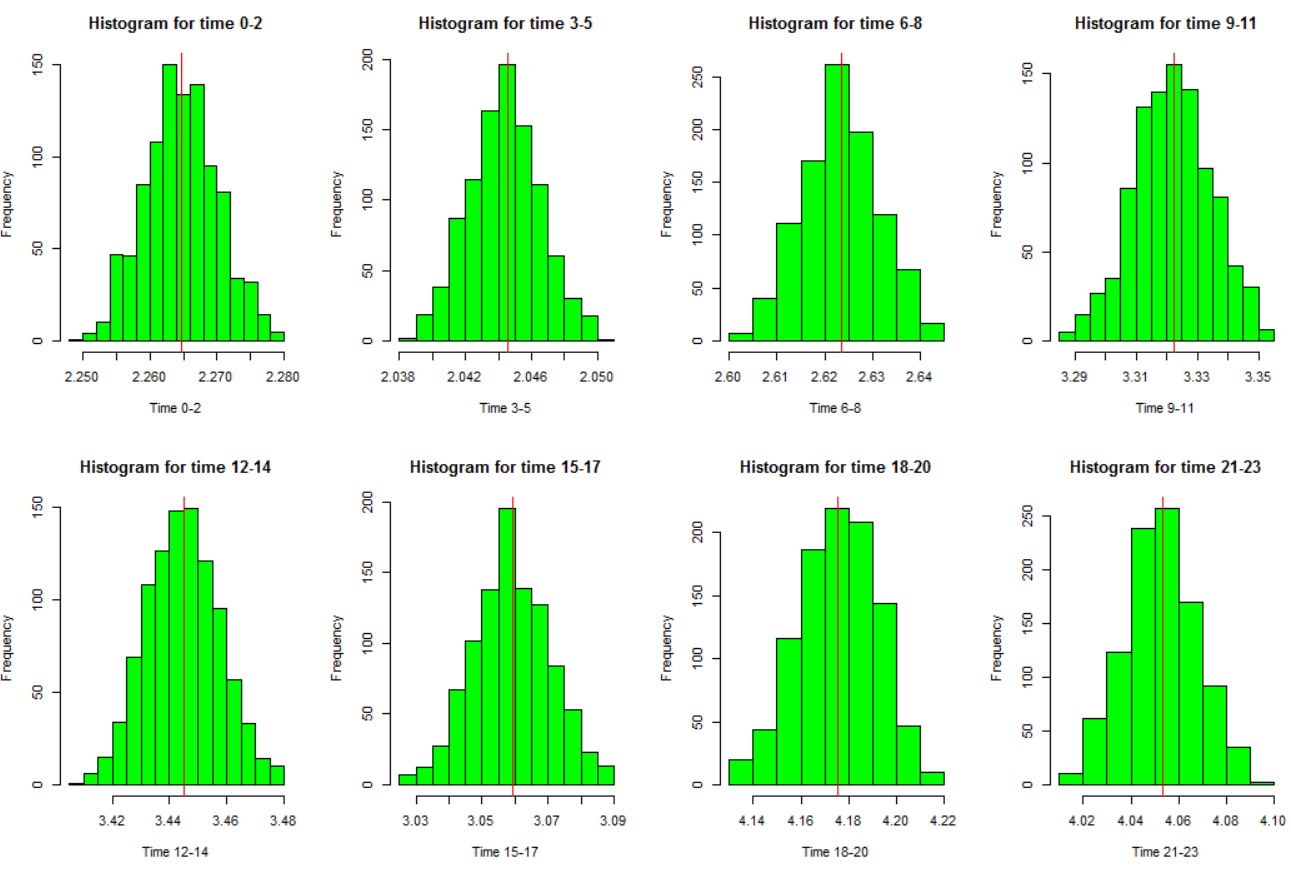}
	\label{fig:fig-real-linear}
\end{figure}

\begin{figure}[!h]
	\centering
	\caption{Histograms of $B=1000$ perturbation-based SGD estimates for the GAS data.}
	\includegraphics[height=0.7\textwidth,width=0.8\textwidth,angle=0] {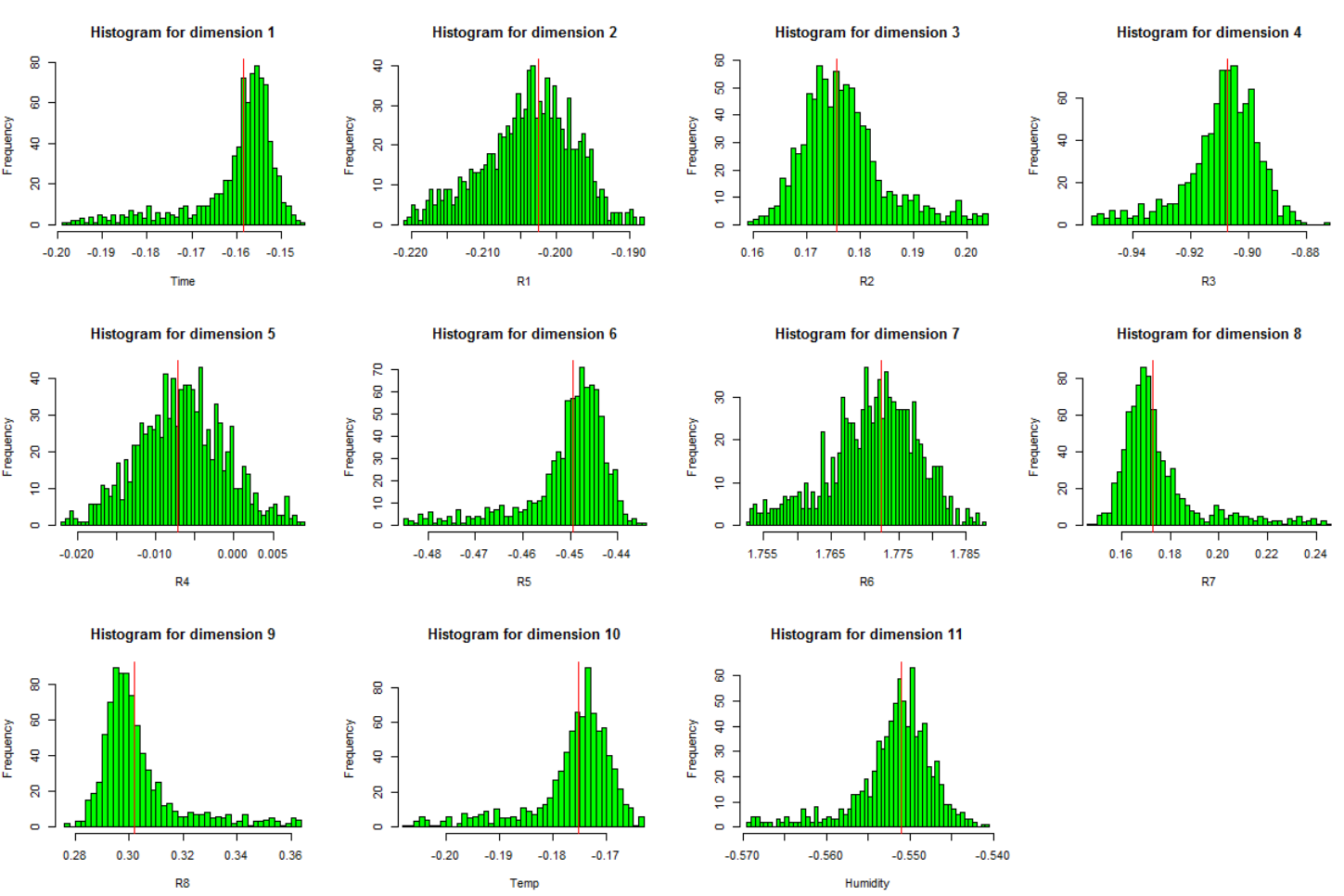}
	\label{fig:fig-real-glm}
\end{figure}

Although standard softwares such as SAS and R can fit linear and logistic regression to such datasets without difficulty, for our illustration purpose, we use the SGD as in Example 1 and 2 to fit linear and logistic regression and use the proposed perturbation-based resampling procedure to construct confidence intervals. The point estimates and 95\% confidence intervals of the coefficients are showed in Table \ref{tab:real-linear} and \ref{tab:real-glm}, for the POWER data and the GAS data, respectively. From Table \ref{tab:real-linear}, we see that the electronic power consumption from kitchen is relatively high in the evening and night. From Table \ref{tab:real-glm}, we see that all the variables but {\it R4} are statistical significantly associated with the response. Further, we display the histogram of $B=1000$ perturbation-based SGD estimates for each coefficient in Figure \ref{fig:fig-real-linear} and \ref{fig:fig-real-glm}for POWER data and the GAS data, respectively. The vertical line in each figure indicates the SGD estimate for one corresponding coefficient. From these figures, we see the the perturbation-based procedure can be used to estimate the whole sampling distribution, not only the standard error, of each SGD estimator.

\begin{table}[!h]
	\centering
	\caption{Point estimates and 95\% confidence intervals of the coefficients for the POWER data.}
	\medskip
	\label{tab:real-linear}
	\begin{tabular}{ccc}
		\hline
		Variable &Point estimate & 95\% CI\\
		\hline\hline
		{\it Time 0-2} & $2.265$ & $(2.254,2.275)$\\
		{\it Time 3-5} & $2.045$ & $(2.040,2.049)$\\
		{\it Time 6-8} & $2.623$ & $(2.608,2.639)$\\
		{\it Time 9-11} & $3.323$ & $(3.298,3.347)$\\
		{\it Time 12-14} & $3.445$ & $(3.420,3.470)$\\
		{\it Time 15-17} & $3.059$ & $(3.037,3.082)$\\
		{\it Time 18-20} & $4.176$ & $(4.143,4.208)$\\
		{\it Time 21-23} & $4.053$ & $(4.024,4.082)$\\
		\hline
	\end{tabular}
\end{table}

\begin{table}[!h]
	\centering
	\caption{Point estimates and 95\% confidence intervals of the coefficients for the GAS data.}
	\medskip
	\label{tab:real-glm}
	\begin{tabular}{ccc}
		\hline
		Variable &Point estimate & 95\% CI\\
		\hline\hline
		{\it Time} & $-0.158$ & $(-0.178, -0.139)$\\
		{\it R1} & $-0.202$ & $(-0.215, -0.190)$\\
		{\it R2} & $0.176$ & $(0.160, 0.191)$\\
		{\it R3} & $-0.907$ & $(-0.932, -0.882)$\\
		{\it R4} & $-0.007$ & $(-0.018, 0.004)$\\
		{\it R5} & $-0.450$ & $(-0.467, -0.432)$\\
		{\it R6} & $1.772$ & $(1.759, 1.785)$\\
		{\it R7} & $0.173$ & $(0.139, 0.207)$\\
		{\it R8} & $0.302$ & $(0.272, 0.332)$\\
		{\it Temperature} & $-0.175$ & $(-0.191, -0.160)$\\
		{\it Humidity} & $-0.551$ & $(-0.560, -0.542)$\\
		\hline
	\end{tabular}
\end{table}

\section{Discussion} \label{discussion}

Online updating is a useful strategy for analyzing big data and streaming data, and recently stochastic gradient decent has become a popular method for doing online updating. Although the asymptotic properties of SGD have been well studied, there is little research on conducting statistical inference based on SGD estimators. In this paper, we propose the perturbation-based resampling procedure, which can be applied to estimate the sampling distribution of an SGD estimator. The offline version of perturbation-based resamping procedure was first proposed by \cite{rubin1981bayesian} and was also discussed in \cite{shao2012jackknife}.

The proposed resampling procedure is in essence an online version of the bootstrap. Recall that the data points, $Z_1, Z_2, \dots, Z_N$, are arriving one at a time and an SGD estimator updates itself from $\wh{\theta}_{n-1}$ to $\wh{\theta}_n$ whenever a new data point $Z_n$ arrives. If we are forced to apply the bootstrap, then we should have many bootstrap samples; the data points of each bootstrap sample, $Z_1^*, Z_2^*, \dots, Z_N^*$, are assumed to be arriving one at a time and the SGD estimator updates itself from $\wh{\theta}^*_{n-1}$ to $\wh{\theta}^*_n$ whenever a new data point $Z_n^*$ arrives. Of course the bootstrap is impractical here because in online updating we cannot obtain all the data points and then generate bootstrap samples. Now if we rearrange hypothetical bootstrap sample $Z_1^*, Z_2^*, \dots, Z_N^*$ as $\{K_1 {\rm \ copies\ } Z_1, K_2 {\rm \ copies\ } Z_2, \dots, K_N {\rm \ copies\ } Z_N\}$, where $K_n$ follows binomial distribution $B(N, 1/N)$, then the SGD estimator updates itself from $\wh{\theta}^*_{n-1}$ to $\wh{\theta}^*_n$ whenever a new batch of data points, $K_n$ copies of $Z_n$, arrives. Noting that binomial distribution $B(N, 1/N)$ approximates to Poisson distribution $P(1)$ as $N\rightarrow\infty$, we see that the aforementioned hypothetical bootstrap is equivalent to our proposed perturbation-based resampling procure with $W_n\sim P(1)$, whose mean and variance are both equal to one.

Finally, the SGD method considered in this paper is actually the explicit SGD, in contract with the implicit SGD considered in \cite{toulis2014asymptotic}. We are working on extending the perturbation-based resampling procedure proposed in this paper for doing statistical inference for the implicit SGD.

\section*{Appendix}
\renewcommand{\theequation}{A.\arabic{equation}}
\setcounter{equation}{0}

For ease exposition of establishing
asymptotic normality of SGD and perturbed SGD estimates, we present the following Proposition 1, adapted from \cite{polyak1992acceleration}, page 841, Theorem 2.
Let $R(\theta): \mathcal{R}^p \rightarrow \mathcal{R}^p$ be some unknown function and $R(\theta)=0$.
The data consist of $Z_n, n=1,2,\dots,$ which are i.i.d. copies of $Z$. Stochastic
gradients are  $\wh{R}(\theta; Z_i)$ and $\mathbb{E}\{\wh{R}(\theta; Z_i)\}=R(\theta)$. With an initial point $\wh{\theta}_0$ and the learning rate  $\gamma_n=\gamma n^{-\alpha}$, the SGD estimate is defined as
\begin{eqnarray}
\wh{\theta}_n=\wh{\theta}_{n-1} - \gamma_n \wh{R}(\wh{\theta}_{n-1}; Z_n)=\wh{\theta}_{n-1}-\gamma_n\left(R(\wh{\theta}_{n-1})-D_n\right),
\end{eqnarray}
where $D_n=R(\wh{\theta}_{n-1})-\wh{R}(\wh{\theta}_{n-1}; Z_n)$, $\gamma>0$ and $0.5<\alpha<1$. The regularity conditions for Proposition 1 are
listed as follows.
\begin{enumerate}
\item[(C1).] There exists a function $V(\theta): \mathcal{R}^p \rightarrow \mathcal{R}$ such that for some $\lambda>0$, $\delta>0$, $l_0>0$, $L_0>0$, and all $\theta, \theta'\in \mathcal{R}^p$, the conditions $V(\theta)\geq \lambda\|\theta\|_2^2$, $\|\nabla V(\theta) - \nabla V(\theta')\|\leq L_0\|\theta-\theta'\|$, $V(0)=0$, $\nabla V(\theta-\theta_0)\trans R(\theta)>0$ for $\theta\neq \theta_0$ hold true. Moreover, $\nabla V(\theta-\theta_0)\trans R(\theta)\geq \l_0 V(\theta)$ for all $\|\theta-\theta_0\|\leq \delta$.

\item[(C2).] There exists a positive definite matrix $S\in \mathcal{R}^{p\times p}$ such that for some $C>0$, $0<\varrho\leq 1$, and $\delta>0$, the condition $\|R(\theta)-S(\theta-\theta_0)\|_2 \leq C\|\theta-\theta_0\|_2^{1+\varrho}$ for all $\|\theta-\theta_0\|\leq \delta$ holds true.

\item[(C3).] $\{D_n\}_{n\geq 1}$ is a martingale difference process, that is, $\mathbb{E}\{D_n|\mathfrak{F}_{n-1}\}=0$ almost surely, and for some $C>0$,
$$\mathbb{E}\left\{\|D_n\|^2_2|\mathfrak{F}_{n-1}\right\}+\|R(\wh{\theta}_{n-1})\|_2^2\leq C\left(1+\|\wh{\theta}_{n-1}\|_2^2\right)\ a.s., $$
for all $n\geq 1$. Consider decomposition $D_n=D_n(0)+E_n(\wh{\theta}_{n-1})$, where $D_n(0)=R(\theta_0)-\wh{R}(\theta_0; Z_n)$ and $E_n(\wh{\theta}_{n-1})=D_n-D_n(0)$. Assume that $\mathbb{E}\{D_n(0)|\mathfrak{F}_{n-1}\}=0$ a.s.,
\begin{eqnarray*}
&\mathbb{E}\{D_n(0)D_n(0)\trans|\mathfrak{F}_{n-1}\}\overset{P}{\rightarrow} V>0, \\
&\sup_{n\geq 1}\mathbb{E}\left\{\|D_n(0)\|_2^2I(|D_n(0)|>\eta)|\mathfrak{F}_{n-1}\right\}\overset{P}{\rightarrow} 0, \mbox{\ as\ }\eta\rightarrow\infty,
\end{eqnarray*}
and there exists $\delta(\Delta)\rightarrow 0$ as $\Delta\rightarrow 0$ such that, for all $n$ large enough,
$$\mathbb{E}\left\{\|E_n(\wh{\theta}_{n-1})\|_2^2|\mathfrak{F}_{n-1}\right\}\leq \delta(\|\wh{\theta}_{n-1}-\theta_0\|_2) \mbox{\ \  a.s..}$$
\end{enumerate}
{\bf Proposition 1}. {\it If Assumptions C1-C3 are satisfied, then (i):	$\ol{\theta}_n\rightarrow\theta_0$,\ \  a.s.; \\
and (ii):
\begin{eqnarray}\label{sum-martingale-0}
\sqrt{n}(\ol{\theta}_n-\theta_0)=\frac{1}{\sqrt{n}}S^{-1}\sum_{i=1}^n D_i+ o_p(1),
\end{eqnarray}
and $\sqrt{n}(\ol{\theta}_n-\theta_0) \Rightarrow \mathcal{N}\left(0, S^{-1}VS^{-1}\right), {\mbox \ in \ distribution.}$ }\\
\\
{\bf Proof of Lemma 1:}\\
By Proposition 1, it is sufficient to show that Assumptions C1-C3 hold under Assumptions A1-A4.
Let $R(\theta)=\nabla L(\theta)$, $\wh{R}(\theta; Z_i)=\nabla l(\theta; Z_i)$, and $V(\theta)=L(\theta_0+\theta)-L(\theta_0)$.
C1 easily follows from Assumptions A1-A3. Note that $V(0)=0$,  Assumption A1 implies that $V(\theta)\geq \lambda\|\theta\|_2^2$, Assumption A2 implies that $\nabla V(\theta-\theta_0)\trans R(\theta)>0$ for $\theta\neq \theta_0$, Assumption A1 implies that $\nabla V(\theta-\theta_0)\trans R(\theta)>0$ for $\theta\neq \theta_0$, and Assumption A3 implies that $\nabla V(\theta-\theta_0)\trans R(\theta)\geq \l_0 V(\theta)$ over some neighborhood of $\theta_0$. Next, we can see that Assumption A3 implies that Assumption C2 holds for $\varrho=1$, and therefore. Finally, noting that $D_n=D_n(0)+E_n(\wh{\theta}_{n-1})$, where $D_n(0)=-\nabla l(\theta_0, Z_n)$ and $E_n(\wh{\theta}_{n-1})=[\nabla L(\wh{\theta}_{n-1})-\nabla L(\theta_0)]-[\nabla l(\wh{\theta}_{n-1}; Z_n)-\nabla l(\theta_0; Z_n)]$, we can see that Assumptions A1 and A4 imply that the conditions in Assumption C3 about $D_n(0)$ and $E_n(\wh{\theta}_{n-1})$ are satisfied. $\Box$
\\
{\bf Proof of Lemma 2:}\\
Define $\Delta=\theta-\theta_0$,  $\wh{\Delta}_{n}=\wh{\theta}_{n}-\theta_0$ and $\ol{\Delta}_n=\ol{\theta}_n-\theta_0$.  Let $R(\Delta)=\mathbb{E}\{\phi(\Delta\trans X)X\}$ and $\wh{R}(\Delta; Z_n)=\psi(\Delta\trans X_n+\varepsilon_n)X_n$. We verify that Assumptions C1-C4 hold if Assumptions B1-B4 are satisfied. First, let $V(\Delta)=\Delta\trans\Delta$. Assumption B2 implies that the conditions about $V(\Delta)$ in Assumption C1 are satisfied. Second, Assumption B3 implies that $\|R(\Delta)-\dot{\phi}(0)G\|_2\leq C\|\Delta\|_2^2$ for some $C>0$ over a neighborhood of $\Delta=0$, $S=\dot{\phi}(0)G$, and $\varrho=1$. Third, to verify Assumption C3, we consider decomposition $D_n=D_n(0)+E_n(\wh{\Delta}_{n-1})$, where $D_n(0)=\psi(\varepsilon_n)X_n$ and $E_n(\wh{\Delta}_{n-1})=[\psi(\wh{\Delta}_{n-1}\trans X_n+\varepsilon_n)-\psi(\varepsilon_n)]-R(\wh{\Delta}_{n-1})$. By Assumptions B1 and B4, we can see that $\mathbb{E}(D_n(0)D_n(0)\trans)=\psi(0)G=V$ and the conditions about $D_n(0)$ in Assumption C3 are satisfied, and by Assumptions B2 and B4, we can see that the condition about $\mathbb{E}(\wh{\Delta}_{n-1})$ in Assumption C3 is satisfied. Lemma 2 then follows  from Proposition 1 and
\begin{eqnarray}\label{sum-martingale-1}
\sqrt{n}(\ol{\theta}_n-\theta_0)=\frac{1}{\sqrt{n}}(\dot{\phi}(0)G)^{-1}\sum_{i=1}^n D_i+ o_p(1).
\end{eqnarray}
{\bf Proof of Theorem 1:}\\
(i). Rewrite $\wh{\theta}^*_n$ as
\begin{eqnarray}\label{SGD-wt-rewrite}
\wh{\theta}^*_n&=&\wh{\theta}^*_{n-1}-\gamma_n\nabla L(\wh{\theta}^*_{n-1})+\gamma_n\left[\nabla L(\wh{\theta}^*_{n-1})-W_n\nabla l(\wh{\theta}^*_{n-1}; Z_n)\right]\nonumber\\
&=& \wh{\theta}^*_{n-1}-\gamma_n\nabla L(\wh{\theta}^*_{n-1})+\gamma_n D^*_n,\label{SGD-wt-rewrite}
\end{eqnarray}
where $D^*_n=\nabla L(\wh{\theta}^*_{n-1})-W_n\nabla l(\wh{\theta}^*_{n-1}; Z_n)$.
 Let $\mathfrak{F}_{n-1}$ denote the Borel field generated by $\{(Z_i, W_i), i\leq n-1\}$.
Since $\mathbb{E}\{W_n|\mathfrak{F}_{n-1}\}=1$ and  $\nabla L(\theta)=\mathbb{E}\{\nabla l(\theta; Z_n)\}$, we have $\mathbb{E}\{D^*_n|\mathfrak{F}_{n-1}\}=0$. Thus $D^*_n$ is a martingale-difference process.
Let $D^*_n(\theta)=\nabla L(\theta)-W_n\nabla l(\theta; Z_n)$. Then $D^*_n(\theta)=D^*_n(\theta_0)+E^*_n(\theta)$, where
\begin{equation}
E^*_n(\theta)=[\nabla L(\theta)-\nabla L(\theta_0)]-W_n[\nabla l(\theta; Z_n)-\nabla l(\theta_0; Z_n)],
\end{equation}
and $\nabla L(\theta_0)=0$. Since $D^*_n(\theta_0)=-W_n\nabla l(\theta_0; Z_n)$, we have $\mathbb{E}\{D^*_n(\theta_0)\}=0$ and
\begin{equation}\label{D-0}
\mathbb{E}\{[D^*_n(\theta_0)][D^*_n(\theta_0)]\trans\}=2S,
\end{equation}\label{E-n}
noting that $\mathbb{E}(W_n^2)=2$. By Cauchy-Schwartz inequality,
\begin{equation}
\mathbb{E}\{\|E_n^*(\theta)\|_2^2\}\leq 2\|\nabla L(\theta)\|_2^2 + 4\mathbb{E}\left\{\|\nabla l(\theta, Z)-\nabla l(\theta_0, Z)\|^2_2\right\}\triangleq \delta(\theta-\theta_0),
\end{equation}
where $\delta(\theta-\theta_0) \rightarrow 0$ as $\theta\rightarrow\theta_0$, using Assumption A4. Also by Cauchy-Schwartz inequality, $\mathbb{E}\{\|E^*_n(\theta)\|_2^2\}\leq 2\|\nabla L(\theta)\|_2^2+2\mathbb{E}\{\|\nabla l(\theta, Z)\|_2^2\}$. Thus, by Assumptions A2 and A4, we have
\begin{equation}\label{D-n}
\mathbb{E}\{\|D_n^*(\theta)\|_2^2\}+\|\nabla L(\theta)\|_2^2\leq 3L_0^2\|\theta-\theta_0\|_2^2 + 2C\|\theta\|_2^2\leq \widetilde{C} (1+\|\theta-\theta_0\|_2^2),
\end{equation}
for some large enough $\widetilde{C}>0$. Combining results (\ref{D-0})-(\ref{D-n}) implies that Assumption C3 holds. Moreover, Assumptions A1 and A2 imply that Assumption C1 holds, and Assumption A3 implies that Assumption C2 holds. By Proposition 1, we have $\wh{\theta}^*_n\rightarrow\theta_0$ almost surely, and
\begin{eqnarray}\label{sum-martingale}
\sqrt{n}(\ol{\theta}_n^*-\theta_0)&=&\frac{1}{\sqrt{n}}S^{-1}\sum_{i=1}^n D_i^*+ o_p(1)\nonumber\\
&=& -\frac{1}{\sqrt{n}}S^{-1}\sum_{i=1}^n W_i \nabla l(\theta_0; Z_i)+\frac{1}{\sqrt{n}}S^{-1}\sum_{i=1}^n E^*_n(\wh{\theta}^*_{n-1})+o_p(1).
\end{eqnarray}
Note that $\mathbb{E}\{\|E_n(\wh{\theta}^*_{n-1})\|_2^2|\mathfrak{F}_{n-1}\}=\delta(\wh{\theta}^*_{n-1}-\theta_0)$, following (\ref{E-n}). Since $\wh{\theta}^*_n\rightarrow\theta_0$ a.s., we have $\delta(\wh{\theta}^*_{n-1}-\theta_0)\rightarrow 0$ a.s. Thus, $S^{-1}\sum_{i=1}^n E_n(\wh{\theta}^*_{n-1})/\sqrt{n}=o_p(1)$. Therefore, by (\ref{sum-martingale}), we have $\sqrt{n}(\ol{\theta}_n^*-\theta_0)= -S^{-1}\sum_{i=1}^n W_i \nabla l(\theta_0; Z_i)/\sqrt{n}$.\\
(ii). Let
\begin{equation}\label{V-n}
V_n=-\frac{1}{\sqrt{n}}S^{-1}(W_i-1)\nabla l(\theta_0, Z_i)=\sum_{i=1}^n (W_i-1)\xi_i/\sqrt{n},
\end{equation}
where $\xi_i=-S^{-1}\nabla l(\theta_0, Z_i)$. By Theorem 1, $\sqrt{n}(\ol{\theta}^*_n-\ol{\theta}_n)=V_n+o_p(1)$. We first show that, for any $\alpha\in U\triangleq\{\alpha\in\mathcal{R}^p: \|\alpha\|_2=1\}$ and  $u\in\mathcal{R}$,
\begin{equation}\label{CLT-alpha}
\mathbb{P}^*\left(\alpha\trans V_n\leq u\right)\rightarrow\Phi(u), {\rm in\ probability},
\end{equation}
where $\Phi(u)$ is the distribution of $\mathcal{N}(0, 1)$. Note that $\frac{1}{n}\sum_{i=1}^n\xi^2_i\rightarrow 0$ in probability and for any $\epsilon>0$
\begin{equation}\label{Lind-cond}
\mathbb{E}^*\left\{(W_1-1)^2\xi_1^2I(|(W_1-1)\xi_1|>\sqrt{n}\epsilon)\right\}\rightarrow 0, {\rm in\ probability.}
\end{equation}
By the central limit theorem, (\ref{CLT-alpha}) holds. By Cantor's diagonal argument \cite{rao1992approximation}, we can show that 
\begin{equation}\label{eq-co-1}
\sup_{v\in\mathcal{R}^p}\left|\mathbb{P}^*\left(\sqrt{n}(\ol{\theta}^*_n-\ol{\theta}_n)\leq v\right)-\mathbb{P}(\zeta\leq v)\right| \rightarrow 0, {\rm in\ probability, }
\end{equation}
where $\zeta\sim\mathcal{N}(0, A^{-1}SA^{-1})$. Similarly, employing the diagonal argument, we also have
\begin{equation}\label{eq-co-2}
\sup_{v\in\mathcal{R}^p}\left|\mathbb{P}\left(\sqrt{n}(\ol{\theta}_n-\theta_0)\leq v\right)-\mathbb{P}(\zeta\leq v)\right| \rightarrow 0.
\end{equation}
This completes the proof. $\Box$\\
\\
{\bf {Proof of Theorem 2:}}\\
(i). The perturbation-resampling  SGD estimator $\wh{\theta}^*_n$ is defined in (\ref{SGD-M-wt}) and $\ol{\theta}_n^*=\sum_{i=1}^n\wh{\theta}_{i}/n$. Let $R(\Delta)=\mathbb{E}\{\phi(\Delta\trans X)X\}$ and notice that $R(0)=0$. Rewrite $\wh{\theta}^*_n$ as
\begin{eqnarray}\label{SGD-wt-rewrite}
\wh{\theta}^*_n&=&\wh{\theta}^*_{n-1}+\gamma_n R(\wh{\theta}^*_{n-1}-\theta_0)+\gamma_n\left[W_n\psi\left((\wh{\theta}^*_{n-1}-\theta_0)\trans X_n + \varepsilon_n\right)X_n-R(\wh{\theta}^*_{n-1}-\theta_0)\right]\nonumber\\
&=& \wh{\theta}^*_{n-1}+\gamma_n R(\wh{\theta}^*_{n-1}-\theta_0)+\gamma_n D^*_n,\label{SGD-M-wt-rewrite}
\end{eqnarray}
where $D^*_n=W_n\psi\left((\wh{\theta}^*_{n-1}-\theta_0)\trans X_n + \varepsilon_n\right)X_n-R(\wh{\theta}^*_{n-1}-\theta_0)$ is a martingale-difference process
since $\mathbb{E}\{W_n|\mathfrak{F}_{n-1}\}=1$ and $\mathbb{E}\{D^*_n|\mathfrak{F}_{n-1}\}=0$.
Let $D^*_n(\Delta)=W_n\psi\left(\Delta\trans X_n + \varepsilon_n\right)X_n-R(\Delta)$ and $D^*_n(\Delta)=D^*_n(0)+E^*_n(\Delta)$, where
\begin{equation}
E^*_n(\Delta)=W_n\left[\psi(\Delta\trans X_n+\varepsilon_n)-\psi(\varepsilon_n)\right]-R(\Delta).
\end{equation}
Since $D^*_n(0)=W_n\psi(\varepsilon_n)X_n$, $\mathbb{E}\{D^*_n(0)\}=0$ and $\mathbb{E}\{[D^*_n(0)][D^*_n(0)]\trans\}=2 \varphi(0)G$. By Assumption B4,  $\mathbb{E}\{\|E^*_n(\Delta)\|_2^2\}\triangleq\delta(\Delta)\rightarrow 0$ as $\Delta\rightarrow 0$. By Assumptions B2 and B4,
\begin{equation}\label{D-M-n}
\mathbb{E}\{\|D_n^*(\Delta)\|_2^2\}+\|R(\Delta)\|_2^2\leq \widetilde{C} (1+\|\Delta\|_2^2),
\end{equation}
for some large enough $\widetilde{C}>0$. Combining the above results, Assumption C3 holds. Moveover, using the similar arguments as those in the proof of Lemma 2, we can verify that, under Assumptions B1-B4, Assumptions  C1-C3 are satisfied. It follows that $\wh{\theta}^*_n\rightarrow\theta_0$ almost surely, and
\begin{eqnarray}\label{sum-M-martingale}
\sqrt{n}(\ol{\theta}_n^*-\theta_0)&=&\frac{1}{\sqrt{n}\dot{\phi}(0)}G^{-1}\sum_{i=1}^n D_i^*+ o_p(1)=\frac{1}{\sqrt{n}\dot{\phi}(0)}G^{-1}\sum_{i=1}^n W_i \psi(\varepsilon_i)\nonumber\\
&& +\frac{1}{\sqrt{n}\dot{\phi}(0)}G^{-1}\sum_{i=1}^n E^*_n(\wh{\theta}^*_{n-1}-\theta_0)+o_p(1).
\end{eqnarray}
 By the definition of $\delta(\Delta)$,  $\mathbb{E}\{\|E_n(\wh{\theta}^*_{n-1}-\Delta)\|_2^2|\mathfrak{F}_{n-1}\}=\delta(\wh{\theta}^*_{n-1}-\theta_0)$.
Since $\wh{\theta}^*_n\rightarrow\theta_0$ a.s., we have $\delta(\wh{\theta}^*_{n-1}-\theta_0)\rightarrow 0$ a.s. Thus, $\sum_{i=1}^n E_n(\wh{\theta}^*_{n-1}-\theta_0)/\sqrt{n}=o_p(1)$. By (\ref{sum-M-martingale}), (i) is proved.\\
(ii). The proof is similar to that in Theorem 1 (ii) and thus omitted. $\Box$

\newpage
\bibliographystyle{agsm}
\bibliography{SGD_RW}

\end{document}